\documentclass{llncs}
\usepackage[utf8]{inputenc}
\usepackage{graphicx}
\title{The Re-Label Method \newline For Data-Centric Machine Learning}
\author{Tong Guo}
\institute{779222056@qq.com}

\begin{document}

\maketitle

\begin{abstract}

In industry deep learning application, our manually labeled data has a certain number of noisy data. To solve this problem and achieve more than 90 score in dev dataset, we present a simple method to find the noisy data and re-label the noisy data by human, given the model predictions as references in human labeling. In this paper, we illustrate our idea for a broad set of deep learning tasks, includes classification, sequence tagging, object detection, sequence generation, click-through rate prediction. The dev dataset evaluation results and human evaluation results verify our idea. 


\end{abstract}

\section{Introduction}

In recent years, deep learning \cite{ref_proc1} model have shown significant improvement on natural language processing(NLP), computer vision and speech processing technologies. However, the model performance is limited by the human labeled data quality. The main reason is that the human labeled data has a certain number of noisy data. Previous work \cite{ref_proc2} has propose the simple idea to find the noisy data and correct the noisy data. In this paper, we first review the way we achieve more than 90 score in classification task, then we further illustrate our idea for sequence tagging, object detection, sequence generation, click-through rate (CTR) prediction.

\section{Background}

In previous work \cite{ref_proc2}, we illustrate our idea in these steps:

1. It is a text classification task. We have a human labeled dataset-v1.

2. We train a BERT-based \cite{ref_proc3} deep model upon dataset-v1 and get model-v1.

3. Using model-v1 to predict the classification labels for dataset-v1. 

4. If the predicted labels of dataset-v1 do not equal to the human labels of dataset-v1, we think they are the noisy data.

5. We label the noisy data again by human, while given the labels of model and last label by human as reference. Then we get dataset-v2 and model-v2.

6. We loop these re-label noisy data steps and get the final dataset and final model.

\section{Same Idea and More Applications}

\begin{figure}
\centering
\includegraphics[width=\textwidth]{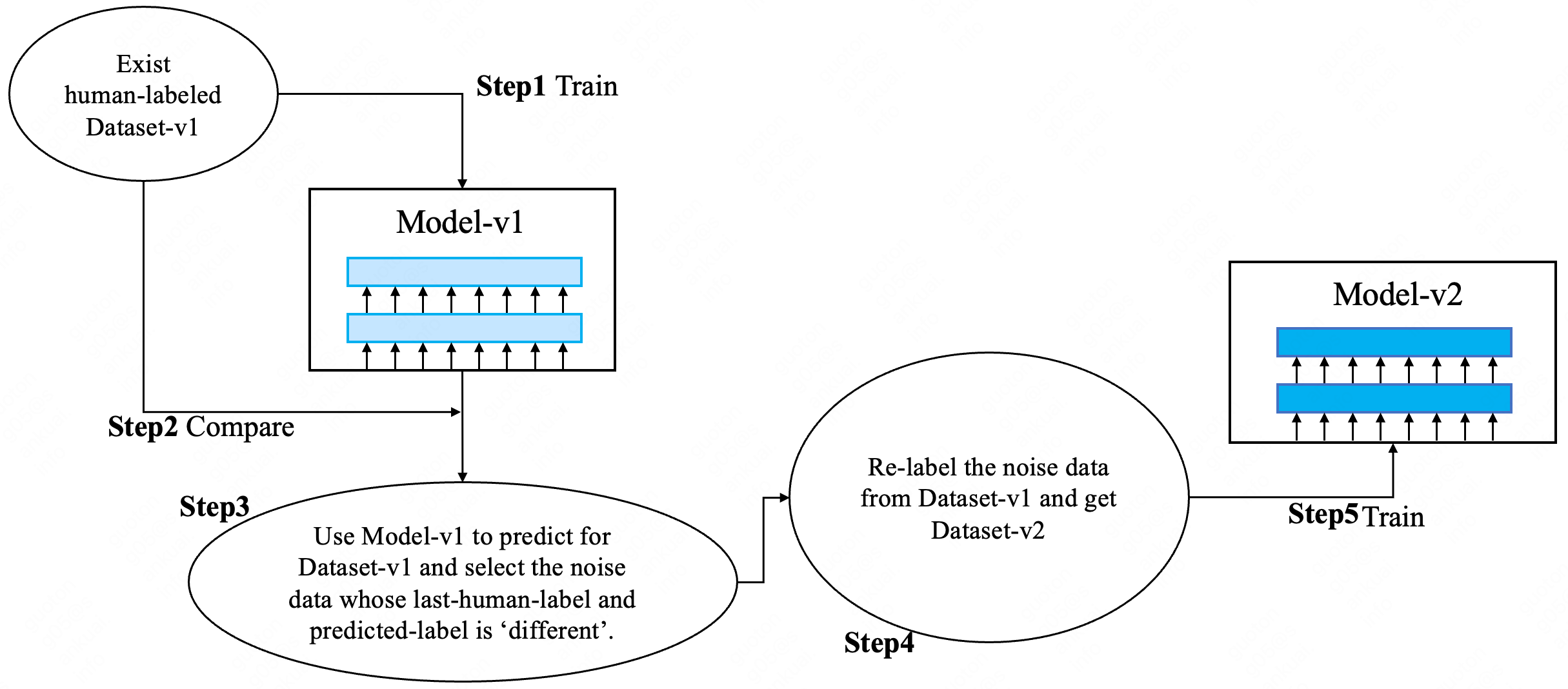}
\caption{Re-label method applying to any machine learning application. } \label{fig1}
\end{figure}

\subsection{sequence tagging}
We take named entity recognition(NER) as example for the sequence tagging like tasks. In NER task, we extract several classes of key phrase from a sentence. Follow our idea, we view each class of NER task as a classification task. Then our steps are:

1. It is a NER task. We have a human labeled dataset-v1.

2. We train a BERT-based \cite{ref_proc3} deep model upon dataset-v1 and get model-v1.

3. Using model-v1 to predict the sequence labels of one class for dataset-v1. 

4. If the predicted labels of dataset-v1 do not equal to the human labels of dataset-v1, we think they are the noisy data.

5. We label the noisy data again by human, while given the labels of model and last label by human as reference. Then we get dataset-v2. In actual operation, if the labeling peoples are sufficient, we label the NER results. If the labeling peoples are not sufficient, the labeling task can be a choice question for labeling peoples. Then we get dataset-v2 and model-v2.

6. We loop these re-label noisy data steps for all the classes of NER and get the final dataset and final model.

\subsection{object detection}

Object detection is a computer vision technique that allows us to identify and locate objects in an image or video. Follow our idea, we view each kind of bounding box as a classification task. Then our steps are:

1. It is a object detection task. We have a human labeled dataset-v1.

2. We train a Swin Transformer\cite{ref_proc4} upon dataset-v1 and get model-v1.

3. Using model-v1 to predict the bounding boxes of one class for dataset-v1. 

4. If the predicted bounding box of dataset-v1 is far from the human labeled bounding box of dataset-v1, we think they are the noisy data.

5. We label the noisy data again by human, while given the bounding boxes of model and last label by human as reference. Then we get dataset-v2 and model-v2.

6. We loop these re-label noisy data steps for all the classes of object detection and get the final dataset and final model.

\subsection{sequence generation}

The key step of our idea is about how to judge the noisy data. For sequence generation, we can use edit distance score or other sequence similarity evaluation method. Then our steps are:

1. We take text generation task as example. We have an origin labeled dataset-v1. The data has input sentence and output sentence.

2. We train a Encoder-Decoder Transformer\cite{ref5} upon dataset-v1 and get model-v1.

3. Using model-v1 to predict the generated sentences for dataset-v1's input sentences. 

4. If the distance score of model-generated sentences is far from the origin output sentences of dataset-v1, we think they are the noisy data. In actual operation, if the model result and the origin output sentence does not have common token, we think the data is a noisy data. 

5. We label the noisy data again by human, while given the generated sentences of model and origin output sentence as reference. In actual operation, it can be a choice question for labeling people. The labeling people chooses the best result from model results and origin output sentence. Then we get dataset-v2 and model-v2.

6. We loop these re-label noisy data steps and get the final dataset and final model.

\subsection{click-through rate prediction}

For CTR task, we use the methods of \cite{ref_proc6} that drop the the noisy data. CTR task is a click-or-not prediction task, we choose a threshold between the predicted score and the 0/1 online label score to judge whether the data is the noisy data. In this way, we could improve the AUC in dev dataset but the online performance should test online.

\section{Experimental Results}

We do the experiments of text classification, NER, text generation to verify our idea. The results is shown in Table 1, Table 2, Table 3. We also do a lot of other classification task and NER task of other dataset. The improvement is also significant and we do not list the detail results.

\begin{table}
\caption{The experiment result of text classification. The model-v1 and model-v2 are corresponding to the previous section}\label{tab1}
\centering
\begin{tabular}{|l|l|l|}
\hline
model & Dev Dataset Accuracy  & Human Evaluation Accuracy \\
\hline
model-v1 & 83.3\% & 88.0\% \\ 
\hline
model-v2 & 91.7\% & 97.2\% \\
\hline
model-v3 & 93.8\% & 97.5\% \\
\hline
\end{tabular}
\end{table}

\begin{table}
\caption{The experiment result of NER. The model-v1 and model-v2 are corresponding to the previous section}\label{tab1}
\centering
\begin{tabular}{|l|l|l|}
\hline
model & Dev Dataset F1  & Human Evaluation Precision \\
\hline
model-v1 & 73.7\% & 86.0\% \\ 
\hline
model-v2 & 88.7\% & 97.0\% \\
\hline
\end{tabular}
\end{table}

\begin{table}
\caption{The experiment result of controllable text generation. The model-v1 and model-v2 are corresponding to the previous section. We evaluate the available rate of generated text}\label{tab1}
\centering
\begin{tabular}{|l|l|}
\hline
model & Human Evaluation \\
\hline
model-v1 & 94.1\% \\ 
\hline
model-v2 & 98.3\% \\
\hline
\end{tabular}
\end{table}

\section{Discussion}

Why re-label method work? Because deep learning is statistic-based. Take classification as example. (In a broad sense, all the machine learning tasks can be viewed as classification.) If there are three \textbf{very similar} data (data-1/data-2/data-3) in total, which labels are class-A/class-A/class-B, Then the trained model will probably predict class-A for data-3. We assume that data-3 is wrong labeled to class-B by human, because more people label its similar data to class-A. If we do not correct data-3, the model prediction for new data that is the most similar to data-3 will be class-B, which is wrong. The new data is more similar to data-3 than data-1/data-2.

The improvement reason is also based on the better and better understanding for the specific task's labeling rule/knowledge of labeling human once by once. Human-in-the-loop here means that in each loop the labeling human leader should learn and summarize the corrected rule/knowledge based on the last loop.

In the labeling details, the to-label dataset can be sorted for people to read and label, in order for better labeling efficiency. Also, sorted dataset can benefit the discriminating for similar data.

The cost of annotation time is crucial for deep learning tasks based on manually labeled data. If there is not enough labeling manpower, we must find ways to reduce the amount of data that needs to be labeled. If we can reduce the amount of data to be labeled to a level that a single programmer can handle, then we do not need an additional labeling team.

\section{Related Work}

\subsection{Pseudo-label-based methods}

The work\cite{ref7} proposes pseudo-label-based method to improve data quality without human re-label, which is different from our method. Our method is to improve the data quality for model of 97\% accuracy/precision/recall/BLEU/AUC by human re-label.

The work\cite{ref8} proposes label-guess-based method to improve data quality without human re-label, which is different from our method. Our method get the guess-label as reference for human re-label.

Pseudo-label \cite{ref19} assigns the model's own hard prediction to unlabeled examples and trains on those predictions. It never checks whether an existing human label is wrong. Re-labeling ImageNet \cite{ref7} replaces a single global label by model-generated multi-labels and localized labels, still without a second human pass. Label-noise correction \cite{ref8} guesses a new class from the neighborhood of a sample and writes that guess back as the label. In our method the model prediction is only a reference. A human decides the final label, and the corrected label is written back into the original training set.

A survey of learning from noisy labels \cite{ref12} reviews robust training under five groups: robust architecture, robust regularization, robust loss design, loss adjustment, and sample selection. These methods change the training procedure so that a wrong label hurts less. They do not ask a human to fix the training set. Our target is a cleaned dataset on which an ordinary model can reach about 97 score, so we correct the data and keep the training algorithm unchanged.

\subsection{Noisy label detection and robust training}

Confident learning \cite{ref14} estimates the joint distribution between the observed noisy label and the latent true label, then prunes or ranks examples. It is a data-centric method \cite{ref15}: the gain comes from label quality rather than a new network. Confident learning removes the suspected errors. We keep the example and send it back to a human, with the model label and the previous human label shown together.

Co-teaching \cite{ref13} trains two networks at the same time. Each network updates on the small-loss examples selected by the other, and the large-loss examples are dropped for that step. Small loss is used as a proxy for a clean label. Dropping is close to the CTR case in Section 3.4, where a threshold on the gap between the predicted score and the online 0/1 label decides that a click log is noise. For classification, sequence tagging, detection, and generation, dropping a disagreed example also drops the input pattern we still need. Re-label keeps the input and only changes the label.

Hendrycks and Gimpel \cite{ref16} use the maximum softmax probability to flag misclassified and out-of-distribution examples. A low top-class score means the model is uncertain. Our noisy-data rule does not need a score threshold: if the predicted label is not equal to the human label, the example is sent to re-label. A close top-1 score and top-2 score is another useful signal that the label is likely to be noisy. We leave the combination of that score gap with our disagreement rule as future work.

Northcutt et al. \cite{ref18} show that label errors are also common in benchmark test sets, on the order of several percent, and that correcting them can change which model ranks higher. Our loop corrects the training set. The same disagreement check can be run on the dev set before the scores in Table 1, Table 2, and Table 3 are reported, otherwise a noisy dev label hides a real gain.

\subsection{Task-specific annotation noise}

For sequence tagging, CrossWeigh \cite{ref17} splits an NER training set into folds, finds sentences on which the fold models disagree with the human tags, and lowers the weight of those sentences. The authors also manually correct about 5.38\% of CoNLL03 test sentences and show that the published F1 near 93\% was measured on a dirty test set. Down-weighting still leaves the wrong span in the training file. In Section 3.1 we treat each entity class as a classification problem. Disagreed spans are re-labeled by human. When annotators are limited, the task is reduced to a choice between the model span and the previous human span.

For object detection, Li et al. \cite{ref20} alternate detector training with automatic correction of both the class label and the box coordinates. The new box and the new class come from the detector itself, and no clean seed set is required. In Section 3.2 a box is treated as noisy when it is far from the human box, and a human draws the box again with the model box as a reference. The model does not overwrite the training file by itself. Fan et al. \cite{ref11} apply a re-label procedure to offline LiDAR 3D detection and report accuracy above the human labels. That result agrees with our claim: moving the score comes from correcting the boxes, not only from changing the detector.

\subsection{ChatGPT}

The work\cite{ref9} of OpenAI also use model predictions for human as references to label. The work\cite{ref9} use the new human-labeled data from model predictions to train the reward model, which is named as reinforcement learning from human feedback(RLHF).

Our work do not have the reward model of RLHF, the differnce between RLHF and our re-label method is that RLHF focus on using the model predictions as training dataset of reward model, and our re-label method focus on correcting the origin training dataset of policy model. Also in RLHF, the new human-labeled data from model predictions do not conflict to our method, because the new data can be simply merged to the whole dataset of policy model, and then to re-label/re-correct by our method.

Also, the correction for policy by the reward model of RLHF is same to correct all the related data/labels that found by the patterns or substrings of the badcases. The detail pattern-based human-correct method is illustrated at \cite{ref10}.

\section{Conclusion}

We argue that the key point to improve the industry deep learning application performance is to correct the noisy data. We propose a simple method to achieve our idea and show the experimental results to verify our idea. 

All data-driven approaches to improving model performance are, in essence, optimizing the training set against the test set. The method in this paper assumes a fixed volume of training data and instead optimizes the correctness of the training-set labels up to 100\%. Once the training set has reached 100\% correctness but test-set performance still fails to improve, the next step is to analyze the bad cases in the test set in order to expand or iteratively refine the input–output pairs of the training set.

\end{document}